\documentclass[letterpaper, 10 pt, conference]{ieeeconf}  

\IEEEoverridecommandlockouts                              

\usepackage{graphicx}
\usepackage{amsmath,amssymb} 
\usepackage{color,xcolor}

\usepackage{xspace}

\usepackage{pifont}
\usepackage{amsmath,amssymb} 
\usepackage{mathtools}
\usepackage{booktabs}
\usepackage{multirow}
\usepackage{subfigure}
\usepackage{array}
\usepackage{makecell}

\newcommand{\ourmethod}{\textsc{DMF-Synch}\xspace} 
\newcommand{\cmark}{\ding{51}}%
\newcommand{\xmark}{\ding{55}}%
\definecolor{arylideyellow}{rgb}{0.91, 0.84, 0.42}
\definecolor{lightblue}{rgb}{0.93,0.95,1.0}

\definecolor{mycolor}{rgb}{0,0,0}

\title{\LARGE \bf Rotation Synchronization  via Deep Matrix Factorization}

 \author{GK Tejus$^{1}$, Giacomo Zara$^{2}$, Paolo Rota$^{2}$, Andrea Fusiello$^{3}$, Elisa Ricci$^{2}$ and Federica Arrigoni$^{4}$
 \thanks{$^{1}$Indian Institute of Technology (ISM) Dhanbad, India}%
 \thanks{$^{2}$University of Trento, Italy}%
 \thanks{$^{3}$University of Udine, Italy}%
\thanks{$^{4}$Politecnico di Milano, Italy}%
 }

\begin{document}

\maketitle
\thispagestyle{empty}
\pagestyle{empty}

\begin{abstract}
In this paper we address the rotation synchronization problem, where the objective is to recover absolute rotations starting from pairwise ones, where the unknowns and the measures are represented as nodes and edges of a graph, respectively. This problem is an essential task  for structure from motion and  simultaneous localization and mapping. We focus on the formulation of synchronization via neural networks, which has only recently begun to be explored in the literature.  Inspired by deep matrix completion, we express rotation synchronization in terms of matrix factorization with a deep neural network. Our formulation exhibits implicit regularization properties and, more importantly, is unsupervised, whereas previous deep approaches are supervised. Our experiments show that we achieve comparable accuracy to the closest competitors in most scenes, while working under weaker assumptions.
\end{abstract}

\section{Introduction}

Rotation synchronization arises in several applications in Robotics and Computer Vision. The most prominent ones are pose graph optimization \cite{CarloneTronAl15}, structure from motion (SfM) \cite{OzyesilVoroninskiAl17} and 3D point cloud registration \cite{GovinduPooja14}. 
Pose graph optimization -- which is at the core of \emph{Simultaneous Localization and Mapping} (SLAM) --  estimates robot poses by a nonlinear optimization  which iteratively refines the solution, starting from an initial guess. Rotation synchronization provides such a starting point.  In much the same way, it produces an initial guess for \emph{Bundle Adjustment}, the nonlinear optimization  underlying SfM \cite{TronZhouAl16}.

Rotation synchronization (a.k.a.\ \emph{multiple rotation averaging} \cite{HartleyTrumpfAl13} or \emph{rotation optimization} \cite{TronZhouAl16}) is an instance of the more general ``synchronization'' problem \cite{ArrigoniFusiello19}, where the task is to find elements of a group starting from pairwise ratios: we are concerned here with the Special Orthogonal Group SO(3), whose elements are rotations in the 3D space. 
Many representations of rotations exist, including unit quaternions, Euler angles and angle-axis. In this paper we consider rotation matrices:
\begin{equation}
    SO(3) = \{ R \in \mathbb{R}^{3 \times 3} \text{ s.t. } R R^{\mathsf{T}} {=} I \ \wedge \ \det({R}){=}1 \}.
\end{equation}

\begin{figure}[b] 
 \centering
 \includegraphics[width=.9\linewidth]{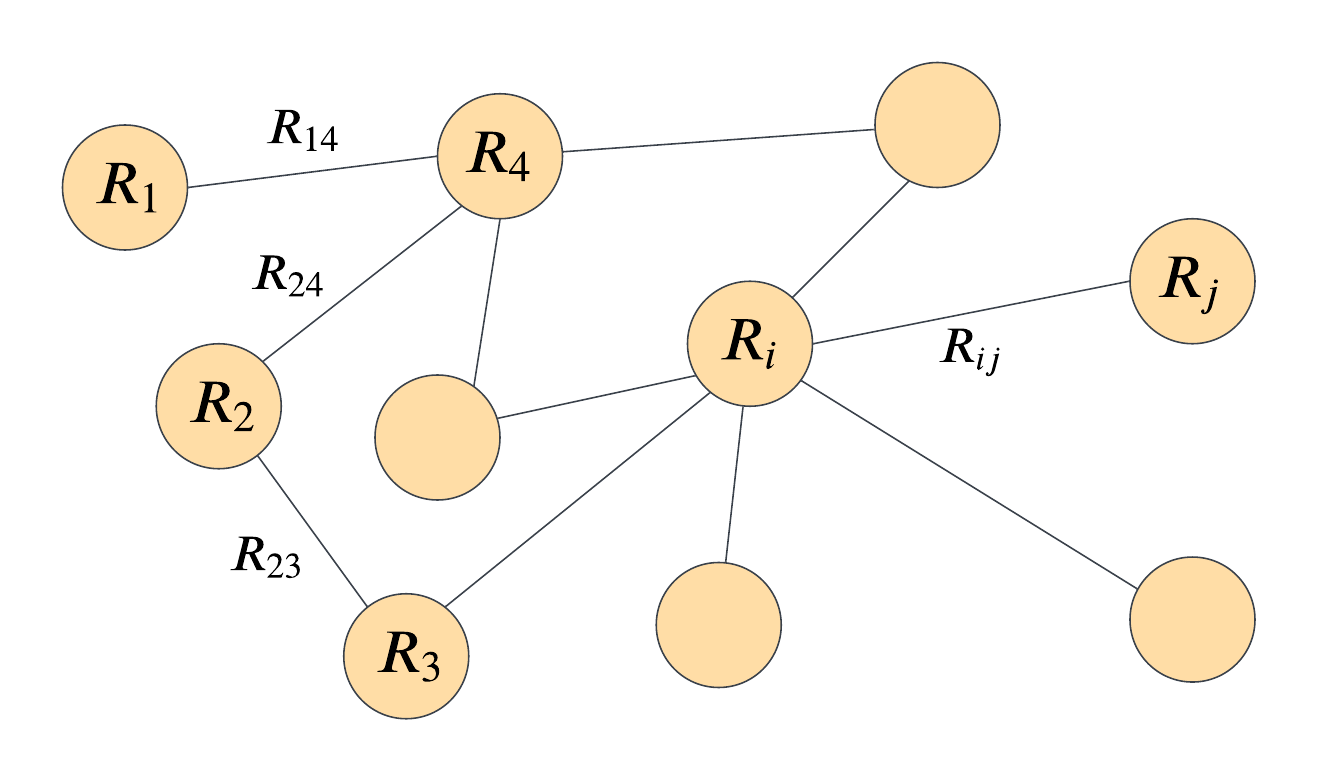}
 \caption{Rotation synchronization problem.  The task of is to recover unknown absolute rotations (on the nodes) starting from measured relative rotations (on the edges).}
 \label{fig_viewgraph}
 \end{figure}

In the \emph{rotation synchronization} problem \cite{Singer11}, the input is a set of relative/pairwise rotations $R_{ij}$ for $(i,j) \in \mathcal{E}$ and the task is to compute a set of global/absolute rotations $R_1, \dots, R_n$ such that the following equation is satisfied
\begin{equation}
    R_{ij} = R_i R_j^{\mathsf{T}} \quad \forall (i,j) \in \mathcal{E}.
    \label{eq_rotation_synch}
\end{equation}
The set $\mathcal{E}$  is usually viewed as the edge set of a simple graph $\mathcal{G}=(\mathcal{V},\mathcal{E})$ with $\mathcal{V}=\{1,2,\dots,n\}$ (see Fig.\ \ref{fig_viewgraph}).

It is well known that the solution to this problem is not unique: indeed, if $R_i$ satisfies \eqref{eq_rotation_synch} for all $i=1, \dots, n$ then also $S_i = R_i Q$ satisfies the same equation, for any fixed $Q \in SO(3)$. Indeed, when computing the product $S_i S_j^{\mathsf{T}} = R_i Q Q^{\mathsf{T}} R_j^{\mathsf{T}}$, the inner term simplifies since $Q$ is a rotation matrix satisfying $Q Q^{\mathsf{T} }= I$. This phenomenon is referred to as ``gauge ambiguity'' and it is usually handled by fixing one rotation to the identity, e.g., $R_1 = I$.

The problem has a solution as soon as the graph is connected. 
The minimal case corresponds to a spanning tree, which has $n-1$ edges. In this scenario, a solution can be found by sequential propagation: the root of the tree is fixed to the identity, and the remaining rotations are computed from $R_i = R_{ij} R_j$, which is equivalent to  \eqref{eq_rotation_synch}.
Synchronization, however, has its strength in exploiting
\emph{redundancy}  to achieve error compensation: indeed, in typical datasets the number of available relative rotations is much larger than the minimum required, i.e., $|\mathcal{E}| \gg n-1$.

Practical rotation synchronization has to deal with some real-world nuisances, such as:  relative rotations may be corrupted by noise with small variance, so that  \eqref{eq_rotation_synch} is not satisfied exactly; a fraction of the relative rotations may be completely wrong (outliers); many of them are typically missing, i.e, 
the graph is not complete.  Therefore, current research focuses on finding principled algorithms that address all of these challenges simultaneously. 

Most methods (e.g., \cite{MartinecPajdla07,HartleyAftabAl11,CrandallOwensAl11,ChatterjeeGovindu17,ErikssonOlssonAl18,BirdalArbelAl20}) follow a traditional perspective, where the problem is formalized as directly optimizing a suitable cost function that measures the discrepancy between the left and right sides in  \eqref{eq_rotation_synch} over all the edges. A popular choice is the following objective\footnote{To clearly distinguish between ground-truth and measured variables, hereafter we use the hat accent to denote the input measures, namely $\widehat{R}_{ij}$ denotes a noisy version of the relative rotation $R_{ij}$.}
\begin{equation}
    \min_{R_1, ... R_n \in SO(3)} \sum_{(i,j) \in \mathcal{E}} d(    \widehat{R}_{ij} , R_i R_j^{\mathsf{T}})^p
\label{eq_cost_general}
\end{equation}
where $d$ is a proper distance defined in SO(3) and typically $p=1$ or $p=2$. See \cite{HartleyTrumpfAl13} for a review of common distances.
As optimizing over rotations is a hard task \cite{HartleyTrumpfAl13,WilsonBindelAl16}, many approaches typically relax the problem by ignoring rotation constraints in \eqref{eq_cost_general} and projecting the found solution onto SO(3) at the end. Some examples include eigendecomposition and semidefinite programming \cite{Singer11,Arie-NachimsonKovalskyAl12}. See Sec.~\ref{sec_related_traditional} for more details and additional examples.

Recently, a few approaches have explored the usage of neural networks for rotation synchronization \cite{PurkaitChiAl20,YangLiAl21,LiLing21}, based on the intuition that 
artificial neural networks are very good in learning patterns from noisy data.
More precisely, given thousands of different problem instances with ground-truth solutions (i.e., graphs with input noisy relative rotations and associated output absolute rotations), the task is to learn a mapping that, at test time, regresses the sought parameters instead of optimizing them.
In particular, a specific type of Graph Neural Networks are used in \cite{PurkaitChiAl20,YangLiAl21,LiLing21}, namely Message Passing Neural Networks \cite{GilmerSchoenholzAl17}, by exploiting the fact that the rotation synchronization task exhibits a graph structure. This way of approaching the problem is still unexplored compared to the traditional perspective, thus motivating the need for further research, as done in our study. However, it should be noted that our method uses neural networks in a different way, as will be clarified later.

\subsection{Contribution}

Despite promising results were shown by \cite{PurkaitChiAl20,YangLiAl21,LiLing21}, such methods suffer from a significant limitation: they are \emph{supervised} approaches as they need a big dataset with known input-output relationship. However, such datasets are hardly available for rotation synchronization. In order to deal with limited dataset availability, the authors of \cite{PurkaitChiAl20}, e.g., pre-train their network on synthetic graphs and fine-tune it on real graphs with a leave-one-out strategy. 

To overcome this drawback, we propose a novel deep method for rotation synchronization  which is \emph{unsupervised}. 
In particular, our starting point is the observation that, since many relative rotations are missing, rotation synchronization can be viewed as a \emph{matrix completion} problem \cite{CandesTao10,ArrigoniRossiAl18}, namely the task of filling missing entries in a big low-rank matrix. In particular, we address the completion task via \emph{deep matrix factorization} \cite{arora2019implicit}, where the sought matrix is written as the product of multiple terms. 
Our choice is motivated by the fact that gradient descent on a deep factorization has some interesting theoretical properties \cite{arora2019implicit}, namely it enhances an ``implicit regularization'' towards low-rank. In addition, as deep matrix factorization can be implemented as a neural network with \emph{linear layers}, our approach offers interpretability \cite{DeHandschutterGillisAl21}. Finally, in order to seek for robustness to outliers, we replace the $ \ell_2$-norm typically used in general deep matrix factorization tasks with the $ \ell_1$-norm, that best suits our rotation synchronization problem.

During the training phase our network minimizes a cost function via backpropagation: the intuition is to make this cost function be the one that solves a matrix completion problem on the instance presented at the input layer. 
Unlike traditional neural models, where to an input corresponds a predicted output, our solution regresses the complete matrix starting from a randomly initialised matrix of equal dimensions. During the training process, at each iteration the input matrix is equal to the transpose of the weight matrix of the first layer. This ploy enables the network to learn continuously, iteration after iteration, until convergence that is when the loss reaches a plateau.
The network is unsupervised, in a sense that it only needs the known parameters of the incomplete target matrix. However the model solves a single problem instance for each training phase (inference is never carried out).

We evaluated our method on standard datasets used in the literature \cite{WilsonSnavely14}. Our results show that we reach comparable accuracy to previous deep methods \cite{PurkaitChiAl20,YangLiAl21,LiLing21} in most cases. Considering that our approach is unsupervised whereas competing methods are supervised, this is an important achievement. Note also that outstanding accuracy is not needed as results are used as an initial guess for a subsequent optimization.

The paper is organized as follows. Section \ref{sec_related} reports previous work on rotation synchronization, Sec.~\ref{sec_method} presents our method and Sec.~\ref{sec_experiments} shows experimental results. Finally, the conclusion is given in Sec.~\ref{sec_conclusion}.

\section{Related Work}
\label{sec_related}

In this section we review previous work on rotation synchronization, by considering separately methods based on explicit/analytical models 
and learned/neural ones. 
Our approach belongs to the latter category.

\subsection{Traditional Methods}
\label{sec_related_traditional}

Rotations have a rich structure that inspired a lot of research in synchronization methods over the years.
Early attempts include methods based on linear least squares, where rotations are represented either as unit quaternions \cite{Govindu01} or rotations matrices \cite{MartinecPajdla07}. 
This approach was revisited in \cite{Singer11,Arie-NachimsonKovalskyAl12} by proposing a compact way to store all rotations simultaneously, and solving the resulting problem via eigendecomposition or semidefinite programming. A semidefinite formulation is also adopted in \cite{ErikssonOlssonAl18,DellaertRosenAl20,ParraChngAl21} with a focus on global optimality.
The authors of \cite{HartleyAftabAl11} address rotation synchronization in an iterative fashion, where each absolute rotation is iteratively updated with the L1 average of its neighbors, computed with the Weiszfeld algorithm.
Another popular approach is Lie-group optimization \cite{ChatterjeeGovindu13}, where the Lie-group structure of rotations is exploited and the problem is iteratively solved in the tangent space.
Other methods include the Levenberg-Marquardt algorithm \cite{CrandallOwensAl11}, low-rank and sparse decomposition \cite{ArrigoniRossiAl18}, Riemannian optimization \cite{BirdalArbelAl20}, message passing \cite{ShiLerman20} and maximum likelihood estimation \cite{MoreiraMarquesAl21}.
Theoretical aspects are also investigated in \cite{HartleyTrumpfAl13,BoumalSingerAl14,WilsonBindelAl16,WilsonBindel20}.

%


\subsection{Methods Based on Neural Networks}
\label{sec_related_neural}
NeuRoRa (Neural Robust Rotation Averaging) \cite{PurkaitChiAl20} is the first solution to rotation synchronization based on neural networks. In particular, the authors propose two networks named CleanNet and FineNet respectively: the former rectifies the input noisy relative rotations while also detecting outliers; the latter fine-tunes an initialization of absolute rotations, which in turn is derived from a spanning tree of the cleaned graph. Both architectures are based on Message Passing Neural Networks (MPNN) \cite{GilmerSchoenholzAl17}. This approach is not end-to-end, for it entails a spanning tree initialization as an intermediate step between the two networks.
Multi-source propagation (MSP) \cite{YangLiAl21} constitutes the first end-to-end approach to rotation synchronization. The focus is on providing an initialization only: the main idea is to initialize absolute rotations according to multiple sources in a differentiable way, in contrast to the common practice of utilizing the spanning tree to initialize rotations according to a single reference node in a top-down manner. Moreover, the image context is also exploited to reduce the impact of outliers. At test time, results are improved with FineNet \cite{PurkaitChiAl20} combined with traditional optimization as in  \eqref{eq_cost_general}.
PoGO-Net \cite{LiLing21} addresses rotation synchronization with a MPNN architecture, similarly to \cite{PurkaitChiAl20,YangLiAl21}. The main innovation is the introduction of a de-noising layer that addresses the outlier removal via an implicit edge-dropping scheme.

%

\section{Proposed Method}
\label{sec_method}

In this section we derive our method: first, we define the notation we use; secondly, we show how to cast rotation synchronization in terms of deep matrix factorization; then, we report some implementation details; finally, we summarize the properties of our approach.

\subsection{Notation}
\label{sec_method_notation}

As done by other authors (e.g., \cite{Arie-NachimsonKovalskyAl12,ErikssonOlssonAl18,DellaertRosenAl20}), we conveniently collect all the relative and absolute rotations in two block-matrices named $Z$ and $X$ respectively:

\begin{equation}
Z = 
\begin{bmatrix}
I & R_{12} & \dots & R_{1n} \\
R_{21} & I & \dots & R_{2n} \\
\dots &  &  & \dots \\
R_{n1} & R_{n2} & \dots & I
\end{bmatrix},
\quad
X = 
\begin{bmatrix}
R_{1} \\
R_{2} \\
\dots \\
R_{n}
\end{bmatrix}
\end{equation}
where $I$ denotes the $3 \times 3$ identity matrix. Note that $Z$ has dimension $3n \times 3n$ whereas $X$ has dimension $3n \times 3$.
It is evident from  \eqref{eq_rotation_synch} that $Z$ can be decomposed as $Z = X X^T$, thus it is symmetric, positive semidefinite and it has low-rank (in fact, it has rank $3$). 
Similarly to $Z$, we define $\widehat{Z}$ as the $3n \times 3n$ block-matrix containing the subset of \emph{observed} pairwise rotations $\widehat{R}_{ij}$ (when available) and unspecified elsewhere.
For example:
\begin{equation}
\widehat{Z} = 
\begin{bmatrix}
I &\widehat{R}_{12} & \dots & ? \\
\widehat{R}_{21} & I & \dots & \widehat{R}_{2n} \\
\dots &  &  & \dots \\
? &\widehat{R}_{n2} & \dots & I
\end{bmatrix}
\label{Z_X_matrix}
\end{equation}
Observe that 
$\widehat{Z} $ is a \emph{partial} matrix as many relative rotations are missing in practice, i.e., the graph is not complete.

\subsection{Formulation}

Recall that $\widehat{Z}$ represents the input to rotation synchronization whereas $X$ represents the desired output.
Our objective here is to recover $Z$ from $\widehat{Z}$, or, in other terms, at the same time \emph{denoising} relative rotations and \emph{completing} missing ones. Indeed, once $Z$ is known, 
the sought absolute rotations $X$  can be straightforwardly recovered from the eigendecomposition of $Z$  \cite{Arie-NachimsonKovalskyAl12}.
In other words, rotation synchronization can be viewed as a \emph{matrix completion} problem \cite{CandesTao10,ArrigoniRossiAl18}.

In general, matrix completion deals with the recovery of missing values of a matrix from a subset of its entries. In this case, the task is to find $W$ such that 
\begin{equation}
    W \odot \Omega  \approx  \widehat{Z} \odot \Omega
\label{eq:matrix_completion}
\end{equation}
where $\odot$ denotes the entry-wise product and $\Omega$ denotes the corresponding binary mask having zeros in correspondence of missing entries\footnote{Observe that $\Omega$ has a block structure in the rotation synchronization problem: each missing rotation/pair results in a $3 \times 3$ zero block.}. $\Omega$ is also known as the \emph{sampling set} in matrix completion literature \cite{CandesRecht09}.
Problem \eqref{eq:matrix_completion} means that we are looking for a \emph{full} matrix $W$ that is as close as possible to the input partial matrix $\widehat{Z}$ \emph{on the set of available entries}.
Such a task is clearly ill-posed, as there are infinitely-many solutions. In order to make the problem tractable, additional assumptions are typically introduced. A popular choice is the \emph{low-rank} assumption \cite{CandesRecht09,CandesTao10}, meaning that $W$ is supposed to belong to a low-dimensional subspace. Note that such condition is satisfied in rotation synchronization, as observed in Sec.~\ref{sec_method_notation}.

Matrix completion is a highly researched topic comprising several solvers (e.g., \cite{KeshavanMontanariAl09,MaGoldfarbAl11,LinChenAl10,DavenportRomberg16}) that work well for a variety of applications, the most prominent one being recommender systems (see the survey \cite{NguyenKimAl19} for additional references). Most of the solutions typically enforce the low-rank assumption via \emph{explicit} constraints such as the nuclear norm regularization \cite{HuZhangAl13}.
Another possibility is to adopt \emph{implicit} regularization, as done in \cite{GunEAL17,arora2019implicit}.
In particular, in \cite{arora2019implicit} the unknown matrix $W$ is expressed as the product of several factors: 
\begin{equation}
    W = W_d \cdot W_{d-1} \dots W_2 \cdot W_1
    \label{eq_deep_factorization}
\end{equation}
and the following problem is addressed
\begin{equation}
 \min_{W_1, \dots, W_d} \frac{1}{|\Omega|} ||( W_d \cdot W_{d-1}  \dots W_2 \cdot W_1) \odot \Omega - \widehat{Z} \odot \Omega ||_F^2
 \label{eq_loss_l2}
\end{equation}
where $|| \cdot ||_F $ denotes the Frobenius norm and $|\Omega|$ denotes the cardinality of the sampling set, i.e., the number of available entries.
Equation \eqref{eq_deep_factorization} is also called \emph{deep matrix factorization} and the number of factors $d$ is called \emph{depth}. It was observed in \cite{arora2019implicit} that in practice, even when the rank is unconstrained, performing gradient descent on a matrix factorization tends to produce low-rank solutions, i.e., the gradient descent algorithm has \emph{implicit regularization} properties. Moreover, increasing depth leads to a more accurate completion when the number of missing entries is high. In particular, the authors of \cite{arora2019implicit} noted that ``adding depth to the matrix factorization leads to more significant gaps between the singular values of the product matrix, i.e.,  to a stronger implicit bias towards low rank''.

In this paper we adopt the deep matrix factorization formulation from \cite{arora2019implicit} and we apply it -- for the first time -- to the rotation synchronization problem. Such a framework is suitable for our task, as it involves a low-rank matrix with missing data, as already observed. However, a practical method  has to deal also with wrong measures, i.e., outliers.
To this end, we substitute the $\ell_2$-norm typically used for deep matrix factorization -- see  \eqref{eq_loss_l2} -- with the $\ell_1$-loss, motivated by the fact that the latter is more robust to outliers than the former. Thus we consider the following optimization problem:
\begin{equation}
 \min_{W_1, \dots, W_d} \frac{1}{|\Omega|} ||( W_d \cdot W_{d-1}  \dots W_2 \cdot W_1) \odot \Omega - \widehat{Z} \odot \Omega ||_{1,1},
 \label{eq_loss_l1}
\end{equation}
where $|| \cdot ||_{1,1}$ is the entry-wise $\ell_1$ norm (in the same way as Frobenius is the entry-wise $\ell_2$ norm).
Following \cite{arora2019implicit}, we cast the above problem 
as training a \emph{linear neural network} with $d$ layers. See the next section for more details. Please note that the matrix completion problem is \emph{not} solved in the forward phase, as one might expect. In fact, it is the backward phase that is equivalent to minimizing the cost function of matrix completion with factorization, as will be clarified later. 
Solving Problem \eqref{eq_loss_l1} will produce an optimal $W = W_d \cdot W_{d-1} \dots W_2 \cdot W_1$, which represents a \emph{denoised and complete} version of the input $\widehat{Z}$, or, equivalently, an approximation of the ground-truth $Z$. Then, in order to recover the sought absolute rotations, we apply eigendecomposition to the optimal $W$, as explained in \cite{Arie-NachimsonKovalskyAl12}.

\begin{figure}[t] 
\centering
 \includegraphics[width=1\linewidth]{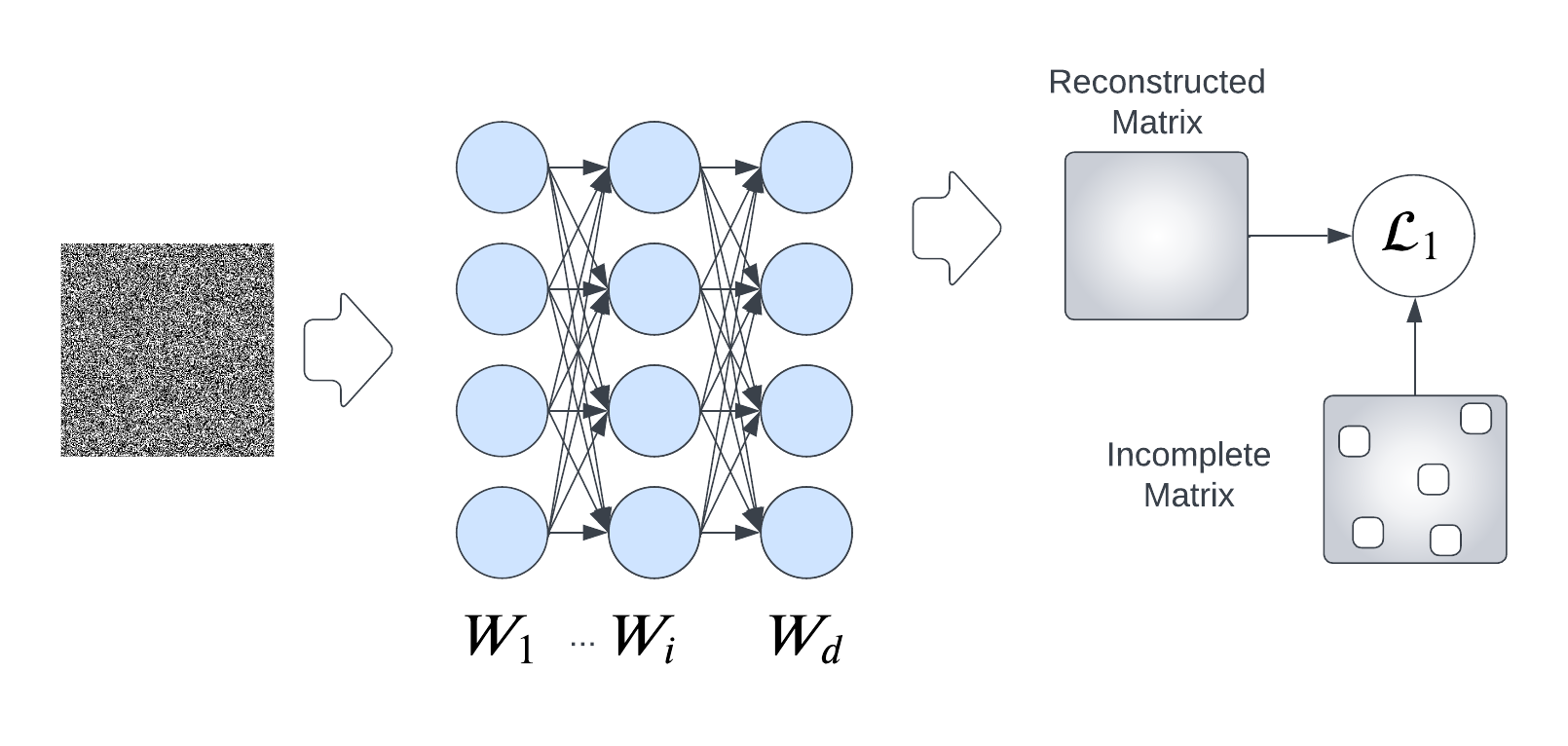} 
 \caption{Architecture, input and output of the network.  Our solution is based on deep matrix factorization, 
 implemented via linear layers (see Sec.~\ref{sec_method}).}
 \label{fig:net_arch}
\end{figure}

\subsection{Implementation Details}

We follow the deep matrix completion approach from~\cite{arora2019implicit}.
More precisely, the network is implemented in PyTorch as a plain sequence of linear layers, whose input and output space is fixed across the architecture and the hidden layers, matching the size of the matrix associated with the problem (see Fig.~\ref{fig:net_arch}). This allows the network to 
address the optimization problem outlined in  \eqref{eq_loss_l1}. In particular, the weight matrices associated to the network are initialized according to a random Gaussian distribution. Given the absence of any activation function, the 
composition of functions carried out with the forward pass across the network can be described by the product reported in  \eqref{eq_deep_factorization}. 
At training time, the prediction of the network is compared with the observed entries in the incomplete matrix, and the parameters are updated by back-propagation according to a standard $\ell_1$ loss. At every iteration, the input matrix is obtained by transposing the weights of the first layer, reducing the input noise variability while the training reaches convergence. At test time, the reconstructed matrix can be compared with the complete ground truth for evaluation. An RTX 2080 Nvidia GPU has been used to carry out our experiments. 
To optimize the network and its parameters, the SGD optimizer is used with a learning rate of $0.3$ and a momentum of $0.9$.

\begin{figure*}[thbp] 
 \centering
\subfigure[100 nodes]
{
 \includegraphics[width=0.42\linewidth]{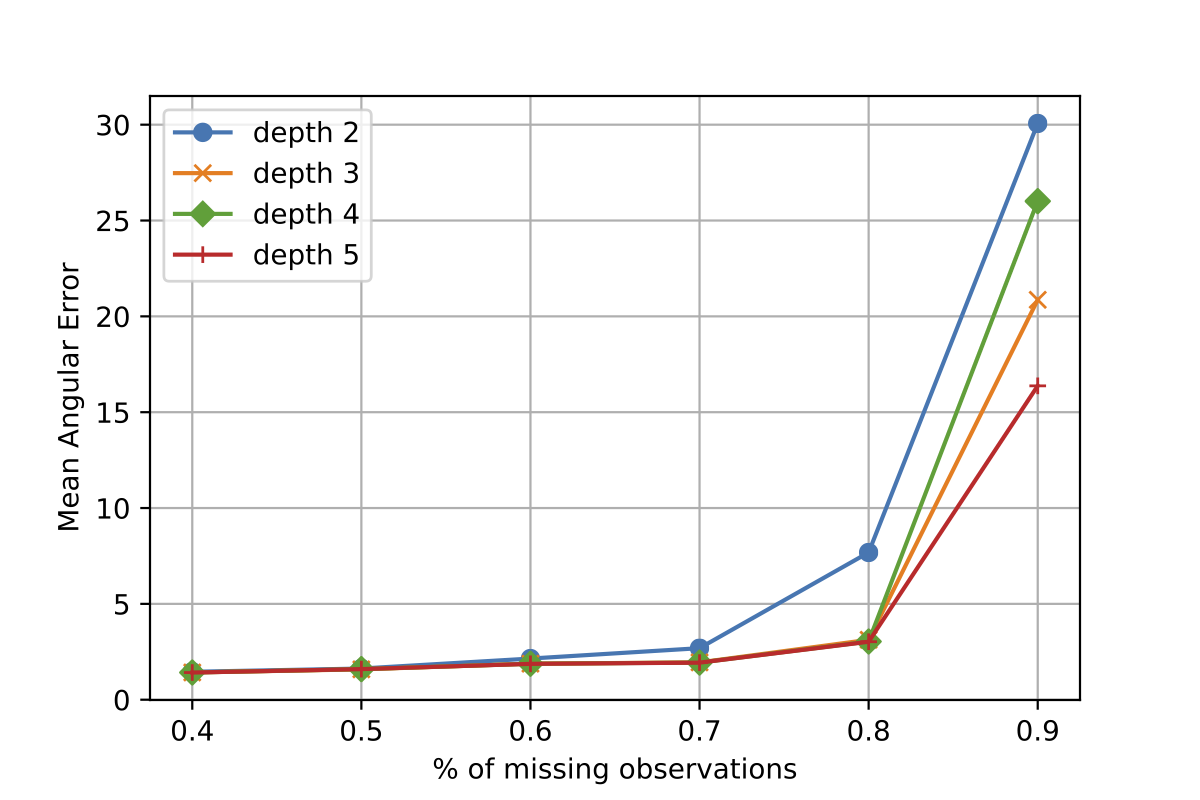} 
} 
\subfigure[200 nodes]
{
 \includegraphics[width=0.42\linewidth]{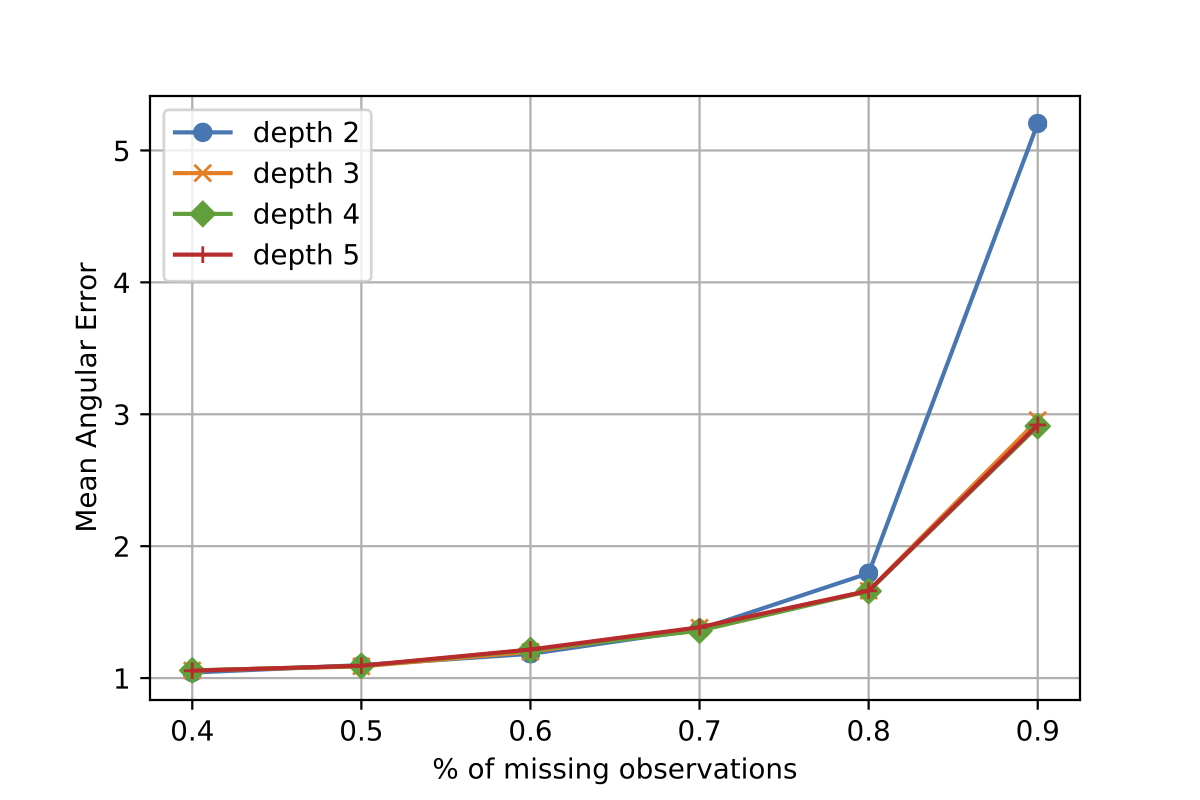} 
} 
 \caption{Mean angular errors [degrees] of our approach versus fraction of missing edges. For each 
 configuration, different choices of depth are analyzed.
 }
 \label{fig:exp_synth}
\end{figure*}

\subsection{Properties}
\label{sec_method_properties}

We summarize here the most relevant properties of our method, henceforth named \ourmethod from ``deep matrix factorization for synchronization''. 

\paragraph*{Unsupervised}
Note that the matrix completion problem is solved for a single matrix at a time\footnote{This is customary in deep matrix completion literature \cite{GunEAL17,arora2019implicit}.}: given relative rotations from a single synchronization problem, we initialize the network with random weights and use backpropagation to minimize the loss function \eqref{eq_loss_l1}. In this context, we may view each entry in the incomplete input matrix as a data point: observed entries constitute our training set, whereas the average reconstruction error over the unobserved entries is the test error, quantifying generalization. In this sense, we may view our method as an \emph{unsupervised} approach.
This is opposed to what is done by previous deep methods for rotation synchronization \cite{PurkaitChiAl20,YangLiAl21,LiLing21}, which can be regarded as \emph{supervised} approaches: they assume as input a dataset with multiple rotation synchronization problems whose solution is known, i.e., the absolute rotations are known for each problem instance comprising a set of relative rotations; hence the model is trained on such dataset and then tested on novel scenes. Following the terminology used in \cite{MoranKoslowskyAl21} for the related problem of projective factorization, we can say that our approach follows a \emph{single scene} setup, whereas previous deep methods \cite{PurkaitChiAl20,YangLiAl21,LiLing21} perform \emph{learning from multiple scenes}. 

\paragraph*{Interpretability} 
Our method is based on deep matrix factorization \cite{arora2019implicit}, which in turn is implemented by a neural network with linear layers. As noted in \cite{DeHandschutterGillisAl21}, the main motivation of deep matrix factorization is to combine both interpretability, as in classical matrix factorization (of which it is an extension), and the extraction of multiple hierarchical features, as allowed by multilayer architectures. In general, linearity has an interpretability power. 

\paragraph*{Implicit Regularization}
Our method is inspired by \cite{arora2019implicit} which proposes an over parametrization of the unknown matrix via deep matrix factorization: this induces a form of implicit regularization when applying gradient-based optimization, namely a bias towards models of low complexity. Although this property is only proved for $\ell_2$-norm in \cite{arora2019implicit}, in Sec.~\ref{sec_experiments} we experimentally confirm this finding also for the $\ell_1$-norm, which is used by our formulation (see \eqref{eq_loss_l1}). Note also that over-parameterization has been recently shown to be beneficial for 3D vision tasks like  camera calibration \cite{SchopsLarssonAl20}.

%

\section{Experiments}
\label{sec_experiments}
We report experiments on both synthetic data, where we validate some  properties of \ourmethod, 
and real data, where we compare our approach to the state of the art. 

\setlength{\fboxsep}{0pt}%
\renewcommand{\arraystretch}{1.2}

\begin{table*}[htbp]
    \centering
    \caption{Median angular errors [degrees] for several methods on the 1DSfM dataset \cite{WilsonSnavely14}.  For each scene, the following statistics are reported: the number of cameras; the percentage of edges in the graph; the hardness measure (the lower the harder) defined in \cite{WilsonBindelAl16}.
    Cases where our method (\ourmethod) is comparable to competing approaches 
    (i.e., the error differs by at most 0.5 degrees) are \colorbox{arylideyellow}{ highlighted}.
    \textcolor{mycolor}{Results of L1-IRLS are copied from \cite{YangLiAl21} and the others are taken from the respective papers.}
    }
    \begin{tabular}{@{}lrrrrrrrrr@{}}
        \toprule
        \multicolumn{4}{l}{\thead{Dataset}} & \multicolumn{4}{c}{\thead{Median Error}} \\
        \cmidrule(l){1-4} \cmidrule(l){5-9}
        \thead{Name} & \thead{\# cam.} & \thead{\% Edges} & \thead{Hardness} & \thead{\ourmethod} & \thead{NeuRoRa} \cite{PurkaitChiAl20} & \thead{MSP} \cite{YangLiAl21} & \thead{PoGO-Net} \cite{LiLing21} & \thead{L1-IRLS} \cite{ChatterjeeGovindu17} \\
        \midrule
        Alamo & 577 & 50 & 0.0017 & 1.2 & \colorbox{arylideyellow}{ 1.2} & \colorbox{arylideyellow}{ 1.1} & \colorbox{ arylideyellow}{ 0.9} & \colorbox{arylideyellow}{ 1.1} \\
        Ellis Island & 227 & 50 & 0.0088 & 0.8 & \colorbox{arylideyellow}{ 0.6} & \colorbox{ arylideyellow}{ 0.8} & \colorbox{ arylideyellow}{ 0.4} & \colorbox{arylideyellow}{0.5} \\
        Gendarmenmarkt & 677 & 18 & 0.0021 & 10.5 & 2.9 & 3.7 & - & { 7.7}\\
        Madrid Metropolis & 341 & 31 & 0.0029 & 2.3 & 1.1 & 1.1 & 1.0 & 1.2 \\
        Montreal Notre Dame & 450 & 47 & 0.0034 & 0.6 & \colorbox{ arylideyellow}{ 0.6} & \colorbox{ arylideyellow}{ 0.5} & \colorbox{ arylideyellow}{ 0.4} & \colorbox{arylideyellow}{0.5} \\
        NYC Library & 332 & 29 & 0.0030 & 1.8 & 1.1 & 1.1 & 0.9 & \colorbox{arylideyellow}{1.3}\\
        Piazza del Popolo & 338 & 40 & 0.0027 & 1.0 & \colorbox{ arylideyellow}{ 0.7} & \colorbox{ arylideyellow}{ 0.8} & \colorbox{ arylideyellow}{ 0.8} & \colorbox{arylideyellow}{0.8} \\
        Roman Forum & 1084 & 11 & 0.0009 & 1.8 & \colorbox{ arylideyellow}{ 1.3} & 1.2 & 0.7 & \colorbox{arylideyellow}{1.5} \\
        Tower of London & 472 & 19 & 0.0021 & 2.7 & 1.4 & 1.3 & 0.4 & \colorbox{arylideyellow}{2.4}\\
        Union Square & 789 & 6 & 0.0006 & 4.4 & 2.0 & 1.9 & 1.3 & \colorbox{arylideyellow}{3.9} \\
        Vienna Cathedral & 836 & 25 & 0.0010 & 1.6 & \colorbox{ arylideyellow}{ 1.5} & \colorbox{ arylideyellow}{ 1.1} & \colorbox{ arylideyellow}{ 1.4} & \colorbox{arylideyellow}{1.2} \\
        Yorkminster & 437 & 27 & 0.0022 & 1.7 & 0.9 & 0.9 & 0.7 & \colorbox{arylideyellow}{1.6} \\
        \bottomrule
    \end{tabular}
    \label{tab_real}
\end{table*}

\subsection{Synthetic Data}
\label{sec_experiments_synth}

In order to analyze the regularization properties of our approach, we consider synthetic datasets generated as follows. First, $n$ ground truth absolute rotations are obtained by random sampling Euler angles.
The graph $\mathcal{G}=(\mathcal{V},\mathcal{E})$ is drawn from
the Erd\H{o}s-R\'enyi model with parameters $(n,p)$, i.e., given a
vertex set $\mathcal{V}=\{1, 2, \dots, n \}$ each edge $(i,j)$ is in
the set $\mathcal{E}$ with probability $p \in [0,1]$, independently of
all other edges. 
Only connected graphs are considered and a fraction of the pairwise rotations -- equal to 0.4 in our experiments -- was highly corrupted, thereby representing outliers. 
The remaining pairwise rotations were corrupted by small multiplicative noise
$
\widehat{R}_{ij} = R_{ij} N_{ij}
$
where, with reference to the angle-axis representation, $N_{ij} \in \text{SO(3)}$ has random axis 
and angle following a Gaussian distribution with zero mean and
standard deviation equal to $5^{\circ}$, thus representing a small perturbation of the identity matrix. %
In order to evaluate the results, we follow standard practices: first, an optimal rotation is computed \cite{HartleyTrumpfAl13} such that the solution gets ``aligned'' with the ground truth rotations, hence becoming comparable; 
then, for each estimated absolute rotation, the distance (e.g., angular) to the ground-truth one can be calculated
and a single statistic (e.g., the mean or median over all rotations) is reported. 

Results are given in Fig.~\ref{fig:exp_synth}, which shows the mean angular errors versus fraction of missing edges in the graph for two different scenarios (100 nodes and 200 nodes). Multiple values for the depth in the factorization are considered, in order to study implicit regularization for increasing depth. Indeed, it is claimed in \cite{arora2019implicit} that -- with reference to  \eqref{eq_loss_l2} -- increasing depth leads to a more accurate solution to matrix completion, especially when the amount of missing data is high. From Fig.~\ref{fig:exp_synth} we can appreciate that the same holds also for the $\ell_1$-norm used by our approach  -- see  \eqref{eq_loss_l1}. In particular, we can observe that using depth 2 results in the worst errors. Hence we have experimentally verified that also the $\ell_1$-norm promotes low-rank solutions when combined with deep matrix factorization and gradient descent, or, in other terms, \ourmethod has implicit regularization properties. Hereafter we will use depth 5 in our experiments.

\textcolor{mycolor}{Finally, it is worth observing that, when using more nodes, it results in a simpler problem. The reason lies in the fact that, for a fixed percentage of edges in the graph, then the effective number of edges does not scale linearly but quadratically with the number of nodes. For example, if the amount of missing data is fixed to $80\%$, then the number of edges is $0.2 \cdot n(n-1)/2$, meaning that there are $990 \approx 9n$ edges with $100$ nodes and $3980 \approx 20n$ edges with $200$ nodes. In the latter case there is significantly more redundancy compared to the minimal case (i.e., a spanning tree with $n-1$ edges), resulting in an easier problem. }

\subsection{Real Data}
\label{sec_experiments_real}

We consider the 1DSfM benchmark \cite{WilsonSnavely14}, which has been widely adopted in the rotation synchronization literature. 
For each scene, both the graph and the associated relative rotations are available, which represents the input to our approach. As done in the literature, the output of Bundler~\cite{SnavelySeitzAl06} is regarded as the ground-truth solution.
\textcolor{mycolor}{In order to make a fair evaluation, we focus on methods based on neural networks, }
namely NeuRoRa \cite{PurkaitChiAl20}, MSP \cite{YangLiAl21} and PoGO-Net \cite{LiLing21}. 
We refer the reader to such papers for a detailed comparison between deep methods and traditional approaches. 
Observe that \ourmethod is \emph{unsupervised} whereas competing methods are supervised, as already mentioned in Sec.~\ref{sec_method_properties}. Hence \ourmethod works under weaker assumptions than its competitors, for it does not require training on a big dataset with known absolute rotations. Moreover, it is worth observing that \ourmethod, NeuRoRa \cite{PurkaitChiAl20} and PoGO-Net \cite{LiLing21} assume as input relative rotations and associated graph only, whereas MSP \cite{YangLiAl21} exploits additional information (namely raw images), as summarized in Tab.~\ref{tab_properties}.
\textcolor{mycolor}{As a reference, we also include in the comparison a traditional approach, namely L1-IRLS \cite{ChatterjeeGovindu17}. }

Results are shown in Tab.~\ref{tab_real}, which reports the median angular error for all the analyzed methods. Some statistics about each scene are also reported, namely the number of cameras (i.e., the number of nodes in the graph representation), the percentage of edges (with respect to a complete graph) and also the hardness measure (the lower the harder) defined in \cite{WilsonBindelAl16}. In general, synchronization problems with few edges are harder to solve, since there is low redundancy, as reflected in the hardness measure. This is also reflected in the errors achieved by all methods, that get slightly higher for the most difficult problems.
Table \ref{tab_real} shows that our approach exhibits good performances, with an angular error lower than two degrees in almost all of the scenes, thus showing the viability of deep matrix factorization for rotation synchronization.

\renewcommand{\arraystretch}{1.4}
\begin{table}[t]
\centering
\caption{
\ourmethod is unsupervised whereas competing methods are supervised. Only MSP \cite{YangLiAl21} uses information coming from raw images. Our method considers weaker assumptions.
 \label{tab_properties}}
{
\begin{tabular}{@{}l c c c c c @{}} 
\toprule
  & \thead{supervised} & \thead{raw images}  \\
\midrule
\ourmethod  & \xmark & \xmark  \\
NeuRoRa \cite{PurkaitChiAl20} & \cmark & \xmark  \\
MSP \cite{YangLiAl21} & \cmark & \cmark \\
PoGO-Net \cite{LiLing21} & \cmark & \xmark   \\ 
\bottomrule
\end{tabular}
}
\end{table}

\textcolor{mycolor}{Our method is on-par with L1-IRLS \cite{ChatterjeeGovindu17}.}
Moreover, \ourmethod reaches comparable accuracy to previous deep methods in many cases: the fact that our approach is not the best is not surprising since we are tackling a more difficult task, as clarified in Tab.~\ref{tab_properties}. Note also that the property of being unsupervised is relevant in practice, given the lack of  datasets for training (as confirmed by previous work \cite{PurkaitChiAl20}).

%

\section{Conclusion}
\label{sec_conclusion}

In this paper we considered the rotation synchronization problem and we showed the feasibility of a novel approach based on deep matrix factorization. Our method follows a different perspective than previous deep techniques: here neural networks are not used for learning from data but as a powerful tool to minimize an objective function. In other terms, our approach is unsupervised and optimizes the network for a single scene at a time. Hence it does not need to acquire the ground truth, which is hard to obtain in real scenarios.

Although performances are only comparable to competing supervised approaches without outperforming them, this is a first attempt to  investigate role of unsupervised neural networks towards rotation averaging, opening the way to new developments and application scenarios.

\color{mycolor}
\small 
\section*{Acknowledgements}
This work was supported by the EU Horizon 2020 Research and Innovation Programme under the project SPRING (No. $871245$).
\normalcolor

\bibliographystyle{plain}

\bibliography{Definitions,NostriNEW,FedeNEW,AndreaNEW,references}

\end{document}